\documentclass[12pt]{article}
\usepackage{times}
\usepackage{txfonts}
\usepackage{bm}
\usepackage{graphicx}
\usepackage{color}

\usepackage{sectsty}
\usepackage{natbib} 
\usepackage[font=footnotesize]{caption}
\usepackage[colorlinks=true, linkcolor=black, citecolor=black, urlcolor=blue]{hyperref}

\renewcommand\sec[1]{\vspace{0.05in}\noindent{{\large\bf{#1}}}
\addtocounter{section}{1}\setcounter{subsection}{0}
}
\newcommand\subsec[1]{\vspace{0.05in} \noindent{\bf{#1}}\addtocounter{subsection}{1}
}

\renewcommand{\b}{\begin{equation}}
\newcommand{\e}{\end{equation}}

\newcommand{\tr}{{\mbox{\scriptsize {\sc T}}}}

\newcommand{\outMy}{\mbox{\scriptsize out}}
\newcommand{\inMy}{\mbox{\scriptsize in}}
\newcommand{\hintMy}{\mbox{\scriptsize hint}}
\newcommand{\JD}{{\mathbf J}^{\mbox{\scriptsize D}}}

\newcommand{\wD}{{\mathbf w}^{\mbox{\scriptsize D}}}

\newcommand{\xD}{{\mathbf x}^{\mbox{\scriptsize D}}}
\newcommand{\zD}{{\mathbf z}^{\mbox{\scriptsize D}}}

\newcommand{\J}{{\mathbf J}}

\newcommand{\w}{{\mathbf w}}
\newcommand{\x}{{\mathbf x}}
\newcommand{\uMy}{{\mathbf u}}

\newcommand{\be}{\begin{equation}}
\newcommand{\ee}{\end{equation}}

\begin{document}

\thispagestyle{empty}
\vspace*{0.5in}
\begin{center}
\begin{Large}
{\bf
full-FORCE:  A Target-Based Method\\
\vspace*{0.05in}
for Training Recurrent Networks\\
}
\end{Large}
\vspace*{0.2in}   
{\bf Brian DePasquale$^{1,4}$, Christopher J. Cueva$^{1}$, Kanaka Rajan$^{4,5}$,\\ G. Sean Escola$^{3}$, L.F. Abbott$^{1,2}$}\\
\vspace*{0.1in}
$^1$Department of Neuroscience\\
Zuckerman Institute\\
Columbia University
New York NY 10027 USA\\
\vspace*{0.2in}
$^2$Department of Physiology and Cellular Biophysics\\
$^3$Department of Psychiatry\\
Columbia University College of Physicians and Surgeons\\
New York NY 10032-2695 USA\\
\vspace*{0.2in}
$^4$Princeton Neuroscience Institute\\
$^5$Lewis-Sigler Institute for Integrative Genomics\\
Princeton University\\
Princeton NJ 08540 USA\\

\vspace*{0.5in}
{\bf Abstract}
\end{center}

Trained recurrent networks are powerful tools for modeling dynamic neural computations. We present a target-based method for modifying the full connectivity matrix of a recurrent network to train it to perform tasks involving temporally complex input/output transformations. The method introduces a second network during training to provide suitable ``target" dynamics useful for performing the task.  Because it exploits the full recurrent connectivity, the method produces networks that perform tasks with fewer neurons and greater noise robustness than traditional least-squares (FORCE) approaches. In addition, we show how introducing additional input signals into the target-generating network, which act as task hints, greatly extends the range of tasks that can be learned and provides control over the complexity and nature of the dynamics of the trained, task-performing network. 

\newpage
\sec{Introduction}

A principle focus in systems and circuits neuroscience is to understand how the neuronal representations of external stimuli and internal intentions generate actions appropriate for a particular task. One fruitful approach for addressing this question is to construct (or ``train'') model neural networks to perform analogous tasks.  Training a network model is done by adjusting its parameters until it generates desired ``target" outputs in response to a given set of inputs.  For layered or recurrent networks, this is difficult because no targets are provided for the ``interior" (also known as hidden) units, those not directly producing the output.  This is the infamous credit-assignment problem.  The most widely used procedure for overcoming this challenge is stochastic gradient-decent using backpropagation~\citep[see, for example,][]{Lecun2015}, which uses sequential differentiation to modify interior connection weights solely on the basis of the discrepancy between the actual and target outputs.  Although enormously successful, this procedure is no panacea, especially for the types of networks and tasks we consider \citep{Sussillo_2014}.  In particular, we construct continuous-time networks that perform tasks where inputs are silent over thousands of model integration time steps.  Using backpropagation through time \citep{Werbos} in such cases requires unfolding the network dynamics into thousands of effective network layers and obtaining gradients during time periods during which, as far as the input is concerned, nothing is happening.  In addition, we are interested in methods that extend to spiking network models \citep{Abbott2016}.   As an alternative to gradient-based approaches, we present a method based on deriving targets not only for the output but also for interior units, and then using a recursive least-squares algorithm \citep{Haykin_2002} to fit the activity of each unit to its target.

\begin{figure}[h!]

	\centerline {\includegraphics[width=6in]{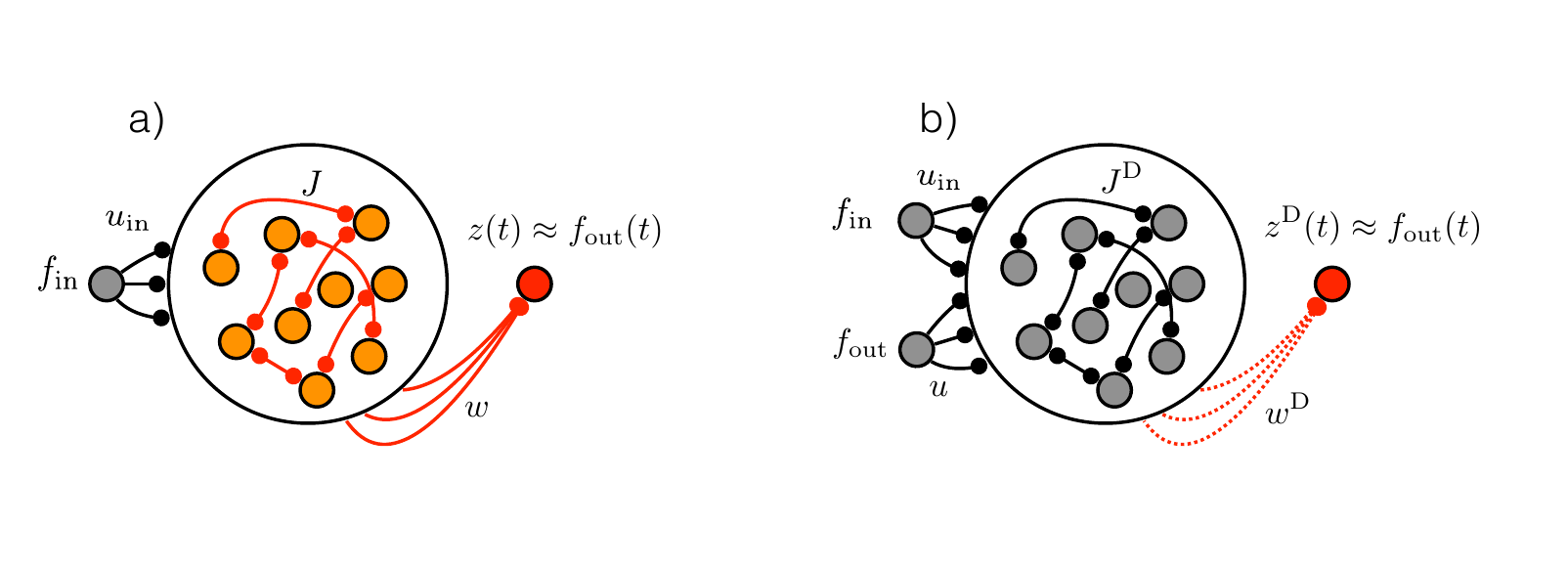}}
	\caption{({\bf a}) Task-performing network. The network receives $f_{\inMy}(t)$ as an input. Training modifies the elements of $\J$ and $\w$ so that the network output $z(t)$ matches a desired target output function $f_{\outMy}(t)$. ({\bf b}) Target-generating network. The network receives $f_{\outMy}(t)$ and $f_{\inMy}(t)$ as inputs. Input connections $\uMy$, $\uMy_{\inMy}$ and recurrent connections $\JD$ are fixed and random. To verify that the dynamics of the target-generating network are sufficient for performing the task, an optional linear projection of the activity, $z^D(t)$, can be constructed by learning output weights $\wD$, but this is a check, not an essential step in the algorithm.}
	
	\label{fig:Figure_1}
	
\end{figure}

Target- rather than backpropagation-based learning has been proposed for feedforward network architectures \citep{Bengio2015}.  Before discussing a number of target-based methods for recurrent networks and presenting ours, we describe the network model we consider and define its variables and parameters.  We use recurrently connected networks of continuous variable ``firing-rate'' units that do not generate action potentials \citep[although see][]{Abbott2016}. The activity of an $N$-unit model network (\hyperref[fig:Figure_1]{Figure~\ref{fig:Figure_1}a)} is described by an $N$-component vector $\x$ that evolves in continuous time according to
\be
	\tau \frac {d\x}{dt} = -\x + \J H (\x) + \uMy_{\inMy} f_{\inMy}(t)\, ,
	\label{eq:net}
\ee

where $\tau$ sets the time scale of the network dynamics (for the examples we show, $\tau = 10$ ms). $H$ is a nonlinear function  that maps the vector of network activities $\x$ into a corresponding vector of  ``firing rates" $H(\x)$ (we use $H(\x) = \tanh(\x)$). $\J$ is an $N\times N$ matrix of recurrent connections between network units.  An input $f_{\inMy}(t)$ is provided to the network units through a vector of input weights $\uMy_{\inMy}$.  The output of the network, $z(t)$, is defined as a sum of unit firing rates weighted by a vector $\w$,
\be
	\label{eq:4output}
	z(t) = \w^{\tr} H(\x(t)) \, .
\ee

Tasks performed by this network are specified by maps between a given input $f_{\inMy}(t)$ and a desired or target output $f_{\outMy}(t)$.  Successful performance of the task requires that $z(t) \approx f_{\outMy}(t)$ to a desired degree of accuracy.  

A network is trained to perform a particular task by adjusting its parameters.  In the most general case, this amounts to adjusting $\J$, $\w$ and $\uMy_{\inMy}$, but we will not consider modifications of $\uMy_{\inMy}$.  Instead, the elements of $\uMy_{\inMy}$ are chosen independently from a uniform distribution between -1 and 1 and left fixed.  The cost function being minimized is
\be
C_{\w} = \left\langle\Big(z(t) - f_{\outMy}(t)\Big)^2\right\rangle \, ,
\label{eq:Cw}
\ee
where the angle brackets denote an average over time during a trial and training examples. The credit assignment problem discussed above arises because we only have a target for the output $z$, namely $f_{\outMy}$, and not for the vector $\x$ of network activities.  Along with backpropagation, a number of approaches have been used to train recurrent networks of this type.  A number of of these involve either ways of circumventing the credit assignment problem or methods for deducing targets for $\x(t)$.  

In liquid- or echo-state networks \citep{Maass,Jaeger}, no internal targets are required because modification of the internal connections $\J$ is avoided entirely.  Instead, modification only involves the output weights $\w$.  In this case, minimizing $C_{\w}$ is a simple least-squares problem with a well-known solution for the optimal $\w$.  The price paid for this simplicity in learning, however, is limited performance from the resulting networks.  

An important next step \citep{Jaeger_Haas_2004} was based on modifying the basic network equation~\ref{eq:net} by feeding the output back into the network through a vector of randomly chosen weights $\uMy$, 
\be
	\tau \frac{d\x}{dt} = -\x + \J H (\x) + \uMy_{\inMy} f_{\inMy}(t)\, + \uMy z(t) \, .
	\label{eq:net2}
\ee
Because $z = \w^{\tr} H(x)$, this is equivalent to replacing the matrix of connections $\J$ in equation~\ref{eq:net} by $\J + \uMy\w^{\tr}$.  Learning is restricted, as in liquid- and echo-state networks, to modification of the output weight vector $\w$ but, because of the additional term $\uMy\w^{\tr}$, this also generates a limited modification in the effective connections of the network.  Modification of the effective connection matrix is limited in two ways; it is low rank (rank one in this example), and it is tied to the modification of the output weight vector.  Nevertheless, when combined with a recursive least-squares algorithm for minimizing $C_{\w}$, this process, known as FORCE learning, is  an effective way to train recurrent networks \citep{Sussillo_Abbott_2009}.  

Although the FORCE approach greatly expands the capabilities of trained recurrent networks,  it does not take advantage of the full recurrent connectivity because of the restrictions on the rank and form of the modifications it implements. Some studies have found that networks trained by FORCE to perform complex ``real-world'' problems such as speech recognition require many more units to match the performance of networks trained by gradient-based methods~\citep{Triefenbach}.  In addition, because of the reliance of FORCE on random connectivity, the activity of the resulting trained networks can be overly complex compared, for example, to experimental recordings~\citep{SussilloChurchland}. 

A suggestion has been made for extending the FORCE algorithm to permit more general internal learning \citep{Sussillo_Abbott_2009,SussilloAbbott2012}. The idea is to use the desired output $f_{\outMy}$ to generate targets for every internal unit in the network. In this approach, the output is not fed back into the network, which is thus governed by equation~\ref{eq:net} not equation~\ref{eq:net2}.  Instead a random vector $\uMy$ is used to generate targets, and $\J$ is adjusted by a recursive least-squares algorithm that minimizes 
\be
C_{\J}^{\mbox{\scriptsize e{\sc F}}}  = \left\langle\Big|\J H(\x(t)) - \uMy f_{\outMy}(t)\Big|^2\right\rangle \, .
\ee
Although this ``extended" FORCE procedure can produce functioning networks, minimizing the above cost function is a rather unusual learning goal, If learning could set $C_{\J}^{\mbox{\scriptsize e{\sc F}}} = 0$, the effective equation of the network would be $\tau d\x/dt = -\x + \uMy f_{\outMy}(t)  + \uMy_{\inMy} f_{\inMy}(t)$.  This equation is incompatible with the output $z(t)$ being equal to $f_{\outMy}(t)$ because $f_{\outMy}(t)$ cannot, in general, be constructed from a low-pass filtered version of itself. Thus, the success of this scheme relies on failing to make $C_{\J}^{\mbox{\scriptsize e{\sc F}}}$ too small, but succeeding enough to assure that the target output is a partial component in the response of each unit.  

\citet{Laje_Buonomano_2013} proposed a scheme that uses a second ``target-generating" network to produce targets for the activities of the network being constructed.  They reasoned that the rich dynamics of a randomly connected network operating in the chaotic regime \citep{Sompolinsky_Crisanti_Sommers_1988} would provide a general basis for many dynamic tasks, but they also noted that the sensitivity of chaotic dynamics to initial conditions and noise ruled out chaotic networks as a source of this basis \citep[see also][]{Thivierge}.  To solve this problem, they used the activities of the chaotic target-generating network, which we denote as $\x^{\mbox{\scriptsize chaos}}(t)$, as targets for the actual network they wished to construct (which we call the ``task-performing" network).  They adjusted $\J$ to minimized the cost function
\be
 C_{\J}^{\mbox{\scriptsize{\sc LB}}}  = \left\langle\Big|H(\x(t)) - H(\x^{\mbox{\scriptsize chaos}}(t))\Big|^2\right\rangle \, .
\ee
After learning, the task-performing network matches the activity of the target-generating network, but it does so in a non-chaotic ``stable" way, alleviating the sensitivity to initial conditions and noise of a truly chaotic system.  Once this stabilization has been achieved, the target output is reproduced as accurately as possible by adjusting the output weights $\w$ to minimize the cost function~\ref{eq:Cw}.

Like the approach of \citet{Laje_Buonomano_2013}, our proposal uses a second network to generate targets, but this target-generating network operates in a driven, non-chaotic regime.  Specifically, the target-generating network is a randomly connected recurrent network driven by external input that is strong enough to suppress chaos \citep{rajansuppression}.  The input to the target-generating network is proportional to the target output $f_{\outMy}(t)$, which gives our approach some similarities to the extended FORCE idea discussed above \citep{Sussillo_Abbott_2009,SussilloAbbott2012}.  However, in contrast to that approach, the goal here is to minimize the cost function as much as possible rather than relying on limited learning.  Because our scheme allows general modifications of the full connection matrix $\J$, but otherwise has similarities to the FORCE approach, we call it full-FORCE.

In the following, we provide a detailed description of full-FORCE and illustrate its operation in a number of examples.  We show that full-FORCE can construct networks that successfully perform tasks with fewer units and more noise resistance than networks constructed using the FORCE algorithm.  Networks constructed by full-FORCE have the desirable property of degrading smoothly as the number of units is decreased or noise is increased.  We discuss the reasons for  these features.  Finally, we note that additional signals can be added to the target-generating network, providing ``hints" to the task-performing network about how the task should be performed \citep{Abu-Mostafa1994,Abu-Mostafa1995}.  Introducing task hints greatly improves network learning and significantly extends the range of tasks that can be learned.  It also allows for the construction of models that span the full range from the more complex dynamics inherited from random recurrent networks to the often simple dynamics of ``hand-build'' solutions. 

\sec{The full-FORCE Algorithm}

As outlined in the introduction, the full-FORCE approach involves two different networks, a target-generating network used only during training and a task-performing network that is the sole final product of the training procedure.  The target-generating network is a random recurrent network that is not modified by learning.  We denote the variables and parameters of the target-generating network by a superscript D standing for ``driven".  This is because the target-generating network receives the target output signal as an input that takes part in driving its activity.

The activities in the $N$-unit target-generating network (\hyperref[fig:Figure_1]{Figure~\ref{fig:Figure_1}b)}, described by the vector $\xD$, are determined by 
\be
	\tau \frac{d\xD}{dt} = -\xD + \JD H (\xD) + \uMy f_{\outMy}(t) + \uMy_{\inMy} f_{\inMy}(t) \, .
	\label{eq:netD}
\ee
The final term in this equation is identical to the input term in equation~\ref{eq:net}.  The connection matrix $\JD$ is random with elements chose i.i.d.\ from a Gaussian distribution with zero mean and variance $g^2/N$.  We generally use a $g$ value slightly greater than 1 (for all of the examples we present here, $g$ = 1.5), meaning that, in the absence of any input, the target-generating network would exhibit chaotic activity \citep{Sompolinsky_Crisanti_Sommers_1988}.  However, importantly, this chaotic activity is suppressed by the two inputs included in equation~\ref{eq:netD}.  The new input term, not present in equation~\ref{eq:net}, delivers the target output $f_{\outMy}(t)$ {\em into} the network through a vector of weights $\uMy$.  The components of $\uMy$, like those of $\uMy_{\inMy}$, are chosen i.i.d.\ from a uniform distribution between -1 and 1.  For all of the examples we present, the numbers of units in the driven and task-generating networks are the same, but we consider the possibility of relaxing this condition in the discussion.

The idea behind the target-generating network described by equation~\ref{eq:netD} is that we want the final task-performing network, described by equation~\ref{eq:net}, to mix the dynamics of its recurrent activity with signals corresponding to the target output $f_{\outMy}(t)$.  If this occurs, it seems likely that a linear readout of the form~\ref{eq:4output} should be able to extract the target output from this mixture.  Through equation~\ref{eq:netD}, we assure that the activities of the target-generating network reflect exactly such a mixture.  For the target-generating network, the presence of signals related to $f_{\outMy}(t)$ is imposed by including this function as a driving input.  This input is not present in equation~\ref{eq:net} describing the task-performing network; it must generate this signal internally.  Thus, learning in full-FORCE amounts to modifying $\J$ so that the task-performing network generates signals related to the target output {\em internally} in a manner that matches the mixing that occurs in the target-generating network when $f_{\outMy}(t)$ is provided {\em externally}.  More explicitly, we want the combination of internal and external signals in the target-generating network, $\JD H (\xD) + \uMy f_{\outMy}(t)$, to be matched by the purely internal signal in the task-performing network, $\J H (\x)$.  This is achieved by adjusting $\J$ to minimize the cost function
\be
C_{\J}^{\mbox{\scriptsize f{\sc F}}}  = \left\langle\Big|\J H(\x(t)) - \JD H(\xD(t)) - \uMy f_{\outMy}(t)\Big|^2\right\rangle \, .
\label{eq:costD}
\ee
This is done, as in FORCE, by using the recursive least-squares algorithm. Before learning, $\J$ is not required to be the same as $\JD$ (for the examples we show, $\J = 0$ before learning).

The primary assumption underlying full-FORCE is that the target output can be extracted by a linear readout from the mixture of internal and external signals within the activities of the target-generating network once they are transferred to the task-performing network.  It is important to note that this assumption can be checked (although this is not a required step in the algorithm) by adding a readout to the target-generating network (\hyperref[fig:Figure_1]{Figure~\ref{fig:Figure_1}b)} and determining whether it can extract the desired output.  In other words, it should be possible to find a set of weights $\wD$ (this can be done by recursive or batch least-squares) for which $(\wD)^{\tr}H(\xD) = \zD(t) \approx f_{\outMy}(t)$ to a desired degree of accuracy.  This readout, combined with the fact that $f_{\outMy}(t)$ is an input to the target-generating network, means that we are requiring the target-generating network to act as an auto-encoder of the signal $f_{\outMy}(t)$ despite the fact that it is a strongly-coupled nonlinear network.

In recurrent networks, learning must not only produce activity that supports the desired network output, it must also suppress any instabilities that might cause small fluctuations in the network activity to grow until they destroy performance.  Indeed, stability is the most critical and difficult part of learning in recurrent networks with non-trivial dynamics \citep{Sussillo_Abbott_2009,Laje_Buonomano_2013,Thivierge}.  As we will show, the stabilization properties of full-FORCE are one of its attractive features.  

\subsec{Differences between full-FORCE and FORCE learning}

The connection matrix $\J$ that minimizes the cost function of equation~\ref{eq:costD} is given by
\be
\J = \left(\JD\left\langle H(\xD)H(\x^{\tr})\right\rangle + \uMy\left\langle f_{\outMy}H(\x^{\tr})\right\rangle \right)\left\langle H(\x)H(\x^{\tr})\right\rangle^{-1} \, .
\label{eq:JLS}
\ee
If the activities of the target-generating and task-performing networks were the same, $\xD = \x$, and the output of the task-performing network was perfect, $\w^{\tr}H(\x) = f_{\outMy}$, equation~\ref{eq:JLS} would reduce to $\J = \JD + \uMy\w^{\tr}$, which is exactly what the FORCE algorithm  would produce if its original recurrent connection matrix was $\JD$.  Of course, we cannot expect the match between the two networks and between the actual and desired outputs to be perfect, but this result might suggest that if these matches are close, $\J$ will be close to $\JD + \uMy\w^{\tr}$ and full-FORCE will be practically equivalent to FORCE.  This expectation is, however, incorrect.

To clarify what is going on, we focus on the first term in equation~\ref{eq:JLS} (the term involving $\JD$; the argument concerning the second term is virtually identical).  This term involves the expression $\left\langle H(\xD)H(\x^{\tr})\right\rangle\left\langle H(\x)H(\x^{\tr})\right\rangle^{-1}$, which obviously reduces to the identity matrix if $\x =\xD$ (and thus the term we are discussing reduces to $\JD$).  Nevertheless, this expression can be quite different from the identity matrix when $\x$ is close to but not equal to $\xD$, which is what happens in practice.  To see why, it is useful to use a basis in which the correlation matrix of task-performing network firing rates is diagonal, which is the principle component (PC) basis.  In this basis, the difference between $\left\langle H(\xD)H(\x^{\tr})\right\rangle\left\langle H(\x)H(\x^{\tr})\right\rangle^{-1}$ and the identity matrix is expressed as a sum over PCs.  The magnitude of the term in this sum corresponding to the n$^{\mbox{\scriptsize th}}$ PC is equal to the projection of $H(\xD) - H(\x)$ onto that PC divided by the square root of its PC eigenvalue $\lambda_n$.  This means that deviations in the projection of the difference between the rates of the target-generating and task-performing networks along the n$^{\mbox{\scriptsize th}}$ PC only have to be of order $\sqrt{\lambda_n}$ to pull $\left\langle H(\xD)H(\x^{\tr})\right\rangle\left\langle H(\x)H(\x^{\tr})\right\rangle^{-1}$ significantly away from the identity matrix.  The spectrum of eigenvalues of the correlation matrix of the activities in the networks we are discussing falls exponentially with increasing $n$ \citep{Rajan2010}.  Therefore, small deviations between $H(\xD)$ and $H(\x)$ can generate large differences between $\J$ and $\JD$.  We illustrate these differences in a later section by examining the eigenvalue spectrum of $\J$ and $\JD + \uMy\w^{\tr}$ for two networks trained for a specific example task.

According to the above analysis, the deviations that cause the results of FORCE and full-FORCE to differ lie in spaces spanned by PCs that account for very little of the variance of the full network activity.  This might make them appear irrelevant. However, as discussed above, stabilizing fluctuations is a major task for learning in recurrent networks, and the most difficult fluctuations to stabilize are those in this space.  The good stabilization properties of full-FORCE are a result of the fact that deviations aligned with PCs that account for small variances modify $\J$ way from $\JD$.

\sec{An Oscillation Task}

\begin{figure}[h!]
	\centerline {\includegraphics[width=5in]{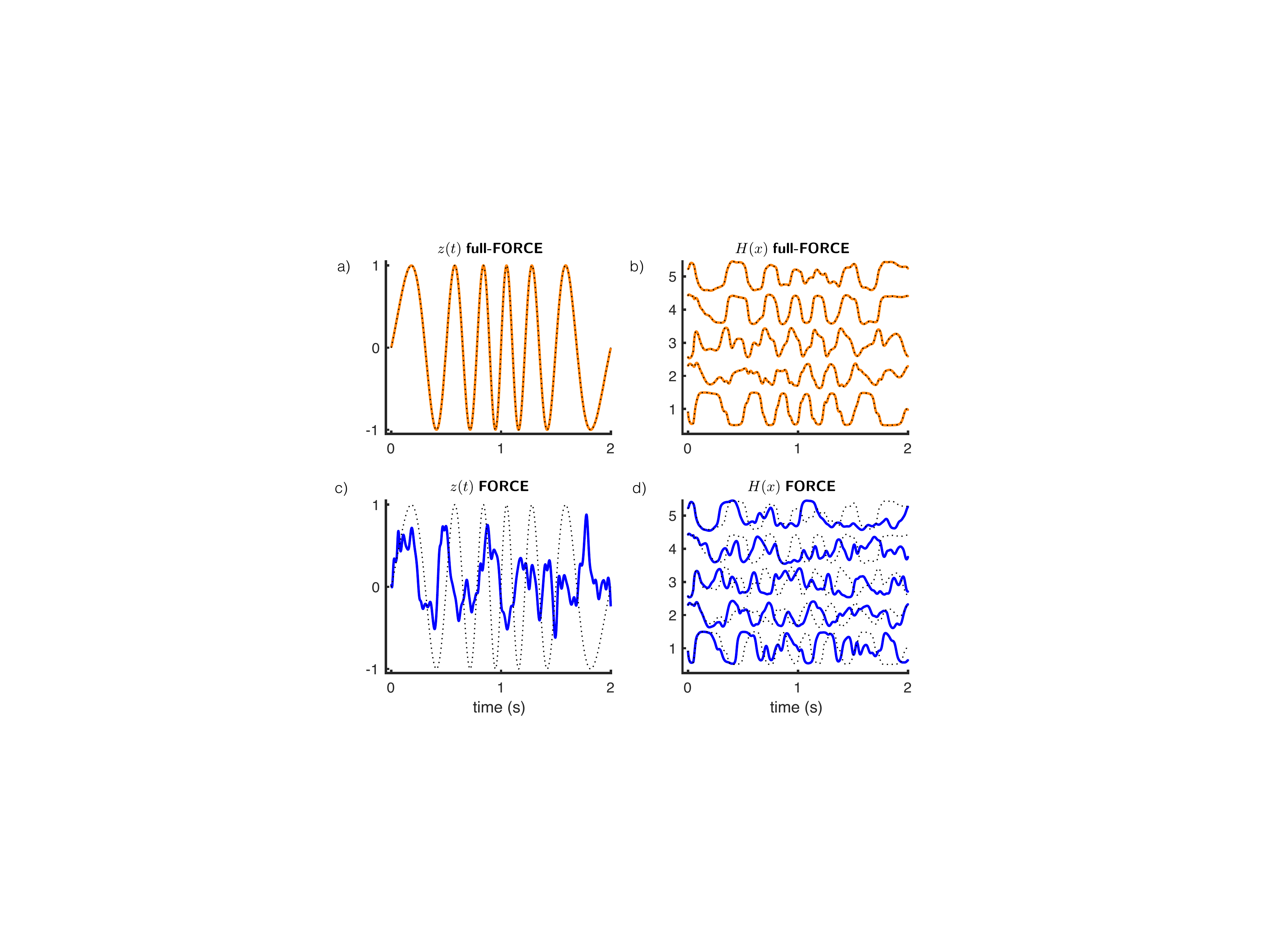}}
	\caption{Example outputs and unit activities from a network of 300 units trained with full-FORCE (a \& b) and FORCE (c \& d) on the oscillation task. ({\bf a}) $f_{\outMy}(t)$ (black dotted) and $z(t)$ (orange) for a network trained with full-FORCE . ({\bf b}) Unit activities (orange) for 5 units from the full-FORCE-trained network compared with the target activities for these units provided by the target-generating network (black dotted).  ({\bf c}) $f_{\outMy}(t)$ (black dotted) and $z(t)$ (blue) for a network trained with FORCE.  ({\bf d}) Unit activities (orange) for 5 units from the FORCE-trained network compared with same target activities shown in b (black dotted).  Because the random matrix used in the FORCE network was $\JD$, activities in a functioning FORCE network should match the  activities from the target-generating network.}
	
	\label{fig:Figure_2}
	
\end{figure}

To illustrate how full-FORCE works and to evaluate its performance, we considered a task in which a brief input pulse triggers a frequency-modulated oscillation in $f_{\outMy}(t)$, and this pattern repeats periodically every 2 s.  Specifically, $f_{\outMy}(t) = \sin(\omega(t)t)$ with $\omega(t)$ increasing linearly from $2\pi$ to $6\pi$ Hz for the first half of the oscillation period, and then the signal is reflected in time around the midpoint of the period, resulting in a periodic accordian-like curve (\hyperref[fig:Figure_2]{Figure~\ref{fig:Figure_2}}a, black dotted trace).  This task was chosen because it is challenging but possible for recurrent network models to achieve.

\begin{figure}[h!]
	\centerline {\includegraphics[width=5in]{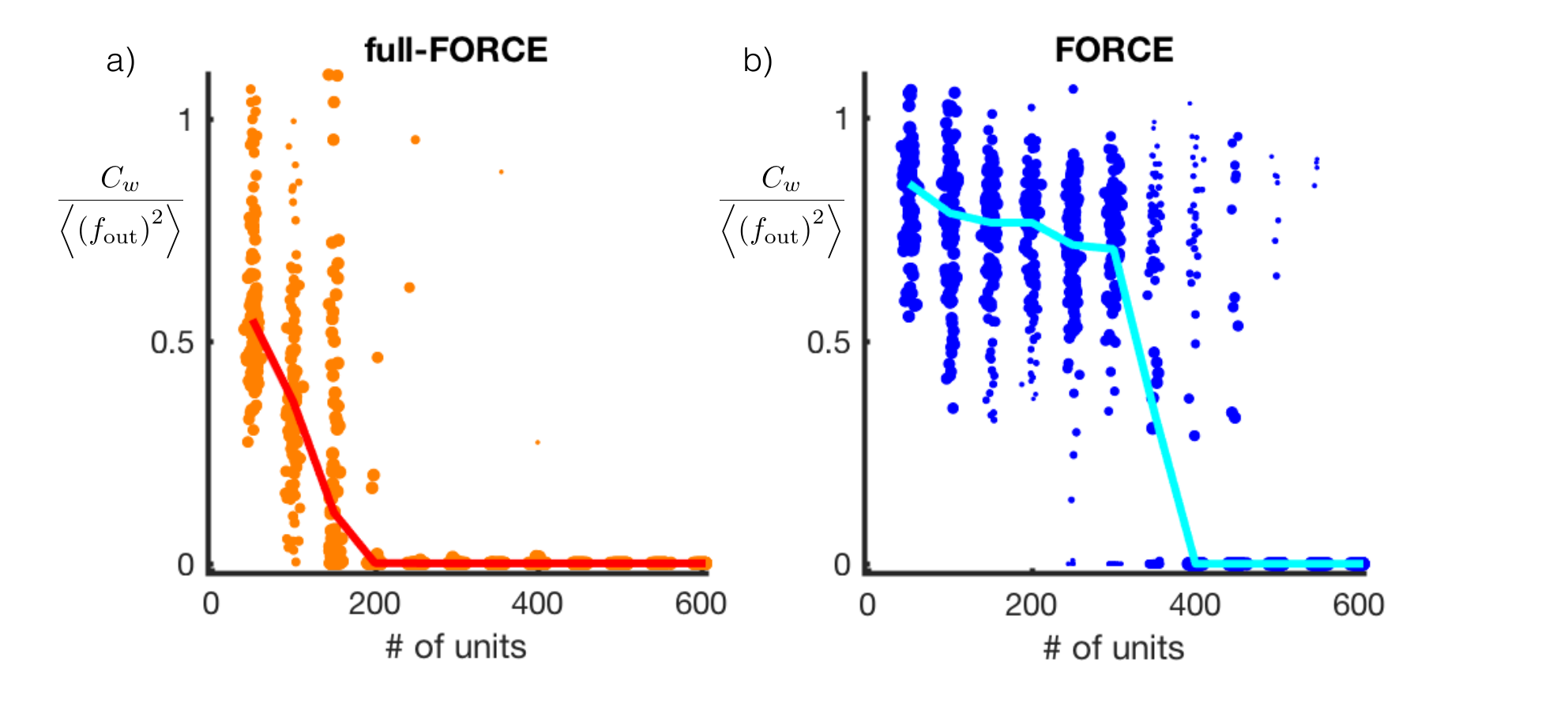}}
	\caption{Normalized test error following network training for full-FORCE ({\bf a}) and FORCE ({\bf b}) as a function of network size. Each dot represents the test error for one random initialization of $\JD$. Test error was computed for 100 random initializations of $\JD$ for each value of $N$. The line indicates the median value across all simulations, and the size of each dot is proportional to the difference of that point from the median value for the specified network size.}
	
	\label{fig:Figure_3}
	
\end{figure}

A full-FORCE-trained network of 300 units can perform this task, whereas a network of this size trained with traditional FORCE cannot (\hyperref[fig:Figure_2]{Figure~\ref{fig:Figure_2}}).  The random connectivity of the FORCE network was set equal to $\JD$, so the target-generating network used for full-FORCE is equivalent to a teacher-forced version of the FORCE network (a network in which the target rather than the actual output is feed back).  The FORCE network fails because its units cannot generate the activity that teacher-forcing demands (\hyperref[fig:Figure_2]{Figure~\ref{fig:Figure_2}}d).  We examined more generally how learning is affected by the number of units for both algorithms (\hyperref[fig:Figure_3]{Figure~\ref{fig:Figure_3}}).  Test error was computed over 50 periods of network output as the optimized value of the cost function of equation~\ref{eq:Cw} divided by the variance of $f_{\outMy}$.

A network trained using full-FORCE can solve the oscillation task reliably using 200 units, whereas FORCE learning required 400 units. In addition, the performance of networks trained by full-FORCE degrades more gradually as $N$ is decreased than for FORCE-trained networks, for which there is a more abrupt transition between good and bad performance.  This is due to the superior stability properties of full-FORCE discussed above.  Networks trained with full-FORCE can be inaccurate but still stable.  With FORCE, instabilities are the limiting factor so networks tend to be either stable and performing well or unstable and, as a result, highly inaccurate.

\begin{figure}[h!]
	\centerline {\includegraphics[width=5in]{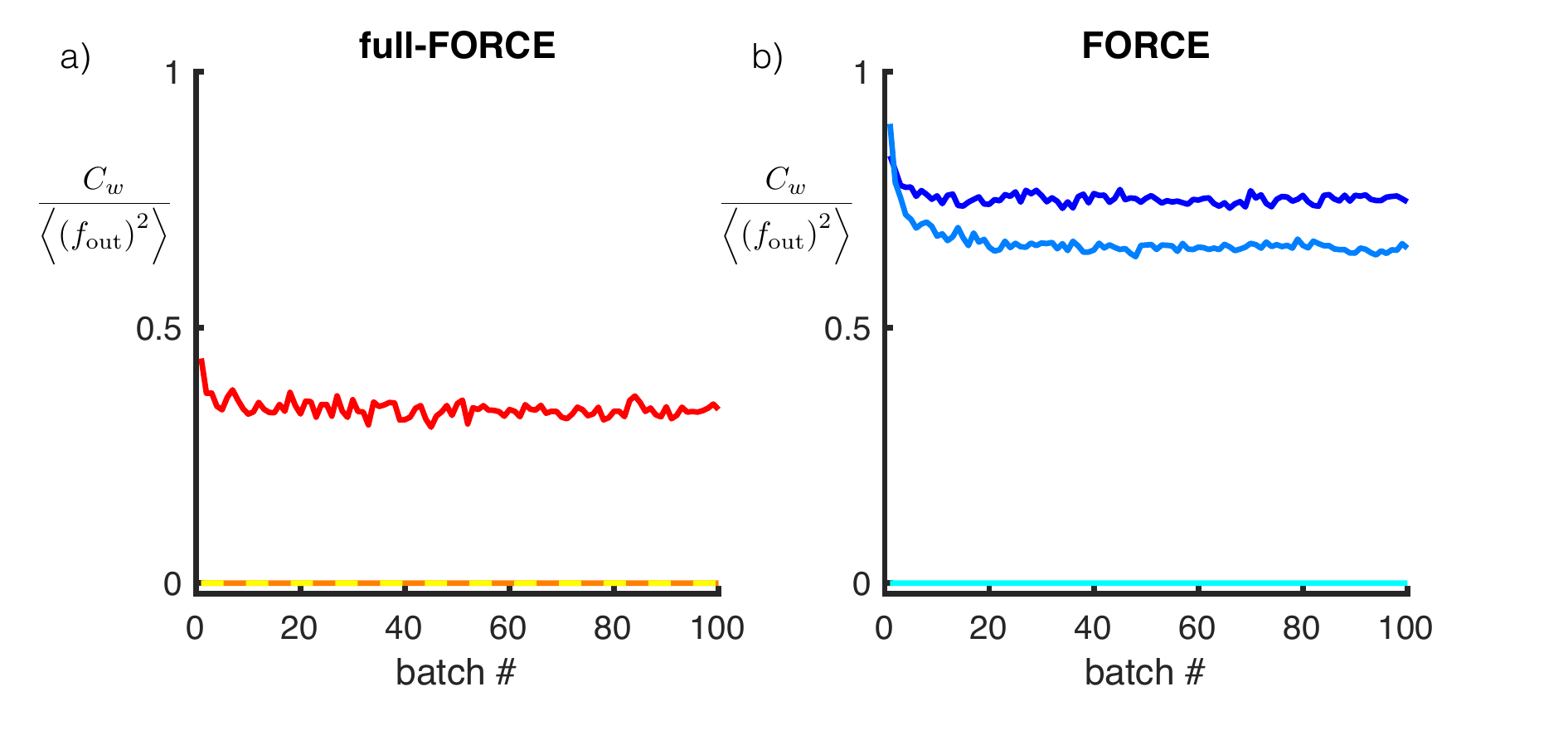}}
	\caption{Median test error for full-FORCE ({\bf a}) and FORCE ({\bf b}) computed across 200 random initializations of $\JD$ for networks trained on the oscillation task. Three different size networks are shown, 100, 200 and 400 units, where larger networks correspond to lighter colors. The horizontal axis shows the number of batches used to train the network, where each batch corresponds to 100 oscillation periods.}
	
	\label{fig:Figure_4}
	
\end{figure}

To check whether long periods of learning could modify these results, we trained networks on this task using either full-FORCE or FORCE with training sets of increasing size (\hyperref[fig:Figure_4]{Figure~\ref{fig:Figure_4}}).  We trained both networks, using up to 100 training batches, where each batch consisted of 100 periods of the oscillation. In cases when training does not result in a successful network, long training does not appreciably improve performance for either algorithm.  For both algorithms there appears to be a minimum required network size independent of training duration, and this size is smaller for full-FORCE than for FORCE.

\subsec{Noise robustness}

We examined the noise robustness of the networks we trained to perform the frequency-modulated oscillator task.  Independent Gaussian white-noise was added as an input to each network unit during both learning and testing. We studied a range of network sizes and five different magnitudes of white noise with diffusion coefficient satisfying $2D$ = $10^{-3}$, $10^{-2}$, 
\begin{figure}[h!]
	\centerline {\includegraphics[width=5in]{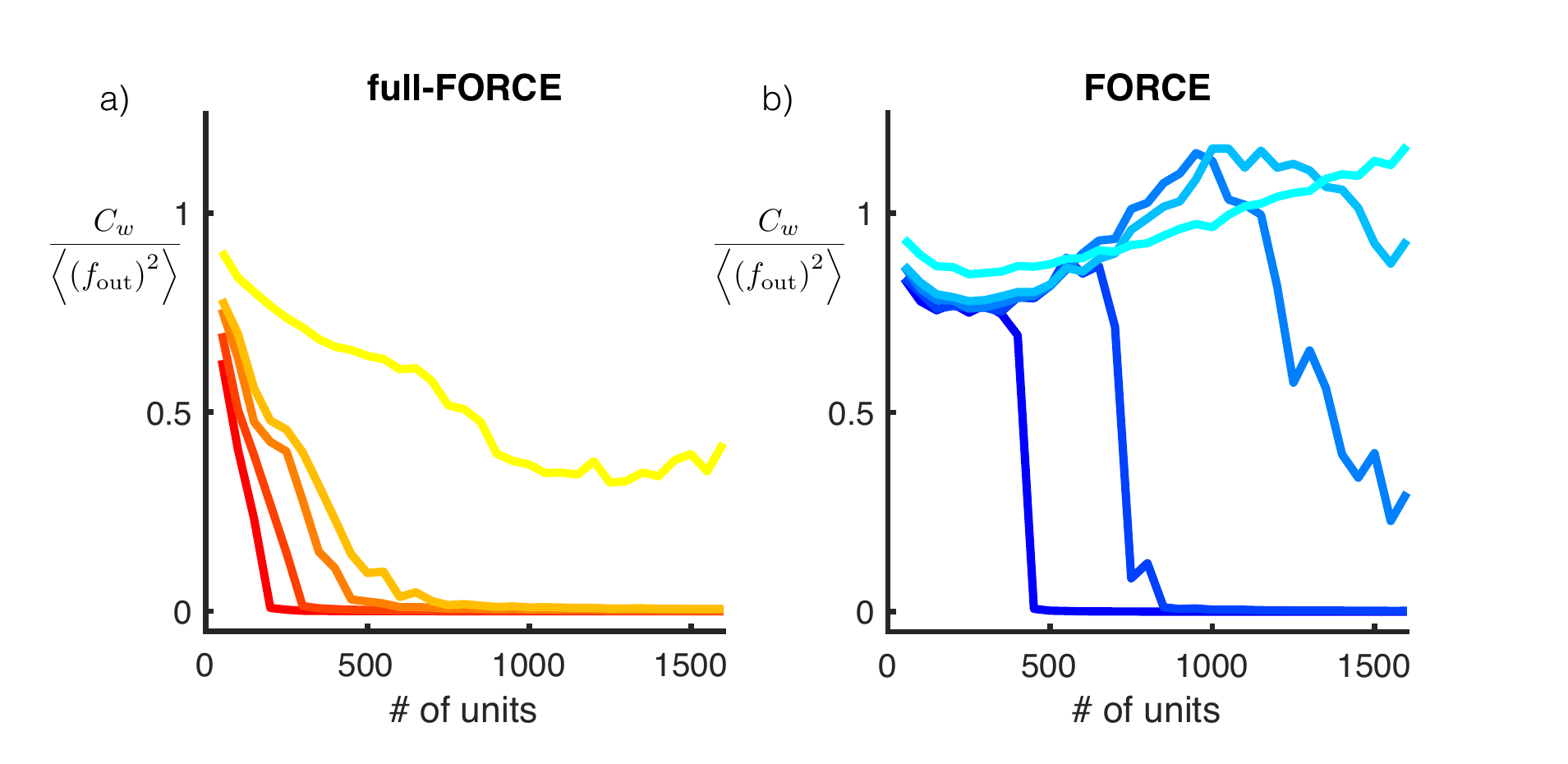}}
	\caption{Median test error for full-FORCE ({\bf a}) and FORCE ({\bf b}) computed across 200 random draws of $\JD$ for various white-noise levels. Increasing noise amplitude corresponds to lighter colors. Levels of noise were determined by $\log(2D)$ = -3, -2, -1, and 0, with D in units of ms$^{-1}$.}
	
	\label{fig:Figure_5}
	
\end{figure}
$10^{-2}$, $10^{-1}$ and $10^{0}$ /ms (\hyperref[fig:Figure_5]{Figure~\ref{fig:Figure_5}}) .  The full-FORCE networks are more resistant to noise and their performance degrades more continuously as a function of  both $N$ (similar to what is seen in \hyperref[fig:Figure_3]{Figure~\ref{fig:Figure_3}}) and the noise level. The robustness to noise for full-FORCE is qualitatively similar to results observed for networks based on stabilized chaotic activity \citep{Laje_Buonomano_2013}.

 
Curiously, over a region of $N$ values when noise is applied, the performance of the FORCE-trained networks \emph{decreases} with increasing $N$  (\hyperref[fig:Figure_5]{Figure~\ref{fig:Figure_5}b}).  This happens because partial learning makes the network more sensitive to noise.  This effect goes away once learning successfully produces the correct output.  This does not occur for full-FORCE, presumably because of its superior stability properties.

\subsec{Eigenvalues of the recurrent connectivity after learning}

\begin{figure}[h!]
	\centerline {\includegraphics[width=4in]{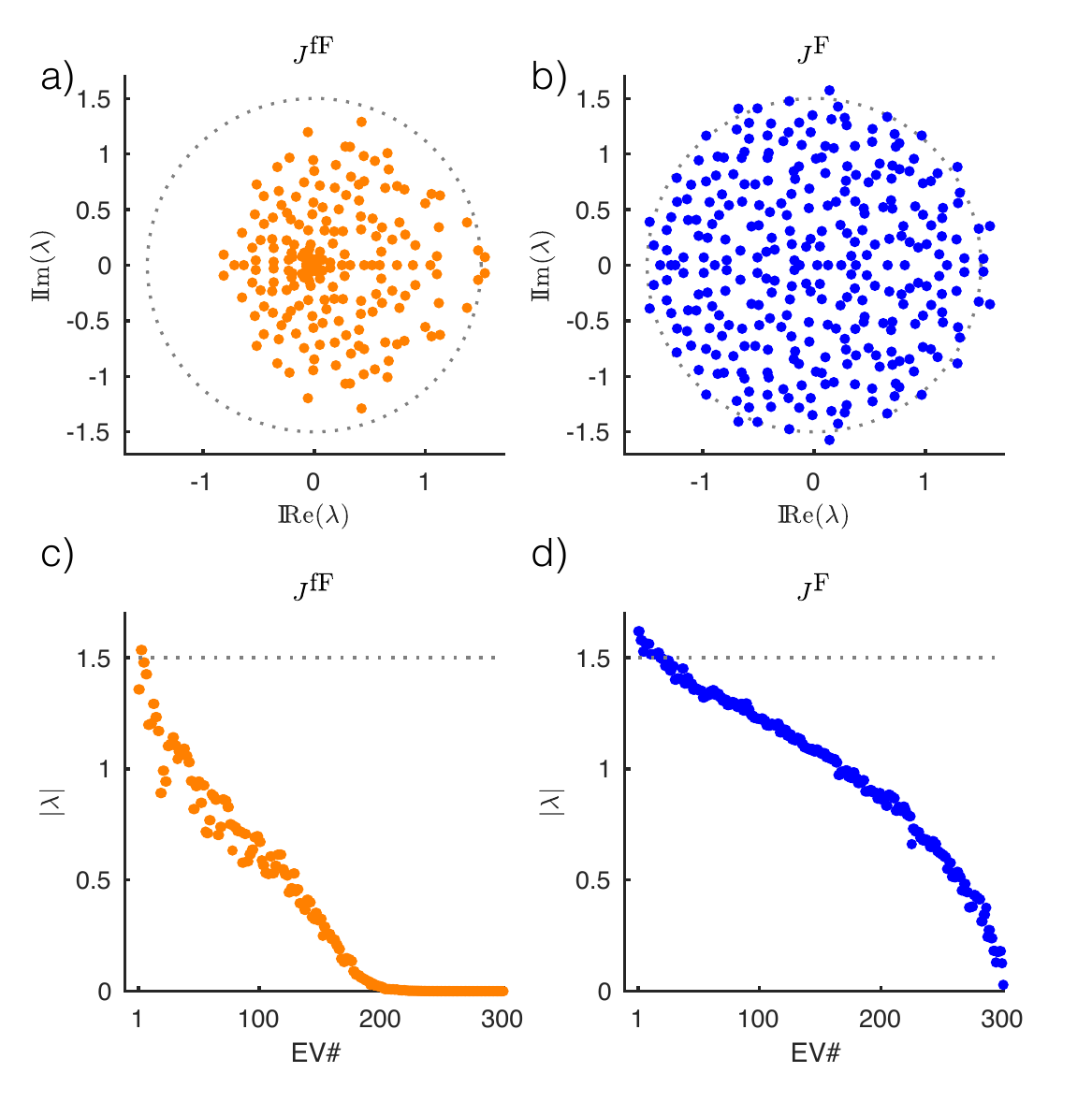}}
	\caption{({\bf a \& \bf b}) Eigenvalues of learned connectivity $\J$ and $\JD + \uMy\w^{\tr}$  for full-FORCE ({\bf a}) and FORCE ({\bf b}) respectively. ({\bf c \& \bf d}) Complex norm of eigenvalues for full-FORCE ({\bf c}) and FORCE ({\bf d}) respectively.  The dotted circle in a and b and dotted line in c and d shows the range of eigenvalues for a large random matrix constructed in the same way as $\JD$. }
	
	\label{fig:Figure_6}
	
\end{figure}

To understand the source of the improved performance of full-FORCE compared to FORCE learning, we examined the eigenvalues of the recurrent connectivity matrix $\J$ after training on the oscillations task and compared them with the eigenvalues of both $\JD$ and $\JD + \uMy\w^{\tr}$, the connectivity relevant for the FORCE-trained network~(\hyperref[fig:Figure_6]{Figure~\ref{fig:Figure_6}}).  The results are for a FORCE and full-FORCE trained network of 300 units that were both successfully trained to perform the oscillations task (for the FORCE network, we used one of the rare cases from 
\hyperref[fig:Figure_3]{Figure~\ref{fig:Figure_3}} where FORCE was successful in training the task).
Mean test error was $1.2 \times 10^{-4}$ and $1.0 \times 10^{-4}$ for FORCE and full-FORCE respectively.  The eigenvalues of the FORCE trained network are only modestly perturbed relative to the eigenvalue distribution of the initial connectivity $\JD$, for which the eigenvalues lie approximately within a circle of radius $g = 1.5$~\citep{Girko}. The eigenvalues of the full-FORCE network are more  significantly changed by learning.  Most noticeably, the range of most of the eigenvalues in the complex plane has been shrunk from the circle of radius 1.5 to a smaller area~(\hyperref[fig:Figure_6]{Figure~\ref{fig:Figure_6}a}).  Equivalently, the norm of most of the eigenvalues is smaller than for the FORCE-trained network~(\hyperref[fig:Figure_6]{Figure~\ref{fig:Figure_6}c}).  The pulling back of eigenvalues that originally had real parts greater than 1 toward real parts closer to 0 reflects the stabilization feature of full-FORCE mentioned previously.  By modifying the full connectivity matrix, full-FORCE allows us to take advantage of the expressivity of a strongly recurrent network while culling back unused and potentially harmful modes.

\sec{Using Additional Input to the Target-Generating Networks as Hints}

The target-generating network of the full-FORCE algorithm uses the ``answer" for the task, $f_{\outMy}$, to suggest a target pattern of activity for the task-performing network.  In the procedure we have described thus far, $f_{\outMy}$ is the only supervisory information provided to the target-generating network.  This raises the question: why not provide more?  In other words, if we are aware of other dynamic features that would help to solve the task or that would lead to a particular type of network solution that we might prefer, why not provide this information to the target-generator and let it be passed on to the task-performing network during training.  This extra information takes the form of ``hints" \citep{Abu-Mostafa1994,Abu-Mostafa1995}.

Information about the desired output is provided as an input to the target-generating network, so it makes sense to incorporate a hint in the same way.  In other words, we add an extra input term to equation~\ref{eq:netD} describing the target-generating network, so now
\be
	\tau \frac{d\xD}{dt} = -\xD + \JD H (\xD) + \uMy f_{\outMy}(t) + \uMy_{\inMy} f_{\inMy}(t) 
	+ \uMy_{\hintMy} f_{\hintMy}(t) \, .
	\label{eq:netHint}
\ee
The weight vector $\uMy_{\hintMy}$ is drawn randomly in the same way as $\uMy$ and $\uMy_{\inMy}$.  The key to the success of this scheme lies in choosing the function $f_{\hintMy}$ appropriately.  There is no fixed procedure for this, it requires insight into the dynamics needed to solve the task.  If the task requires integrating the input, for example, $f_{\hintMy}$ might be the needed integral.  If the task requires timing, as in one of the examples to follow, $f_{\hintMy}$ might be a ramping signal.  The point is that whatever $f_{\hintMy}$ is, this signal will be internally generated by the task-performing network after successful training, and it can then be used to help solve the task.  The cost functions used during training with hints are equation~\ref{eq:Cw} and
\be
C_{\J}^{\mbox{\scriptsize fFhint}}  = \left\langle\Big|\J H(\x(t)) - \JD H(\xD(t)) - \uMy f_{\outMy}(t)
-  \uMy_{\hintMy} f_{\hintMy}(t) \Big|^2\right\rangle \, .
\label{eq:costHint}
\ee

To illustrate the utility of task hints, we present examples of networks inspired by two standard tasks used in systems neuroscience: interval matching and delayed comparison. In both cases, including task hints drastically improves the performance of the trained network. We also show that using hints allows trained networks to extrapolate solutions of a task beyond the domain on which they were trained.  It is worth emphasizing that the hint input is only fed to the target-generating network and only used during training.  The hint is not provided to the task-performing network at any time.

\subsec{Interval matching task}

\begin{figure}[h!]
	\centerline {\includegraphics[width=4in]{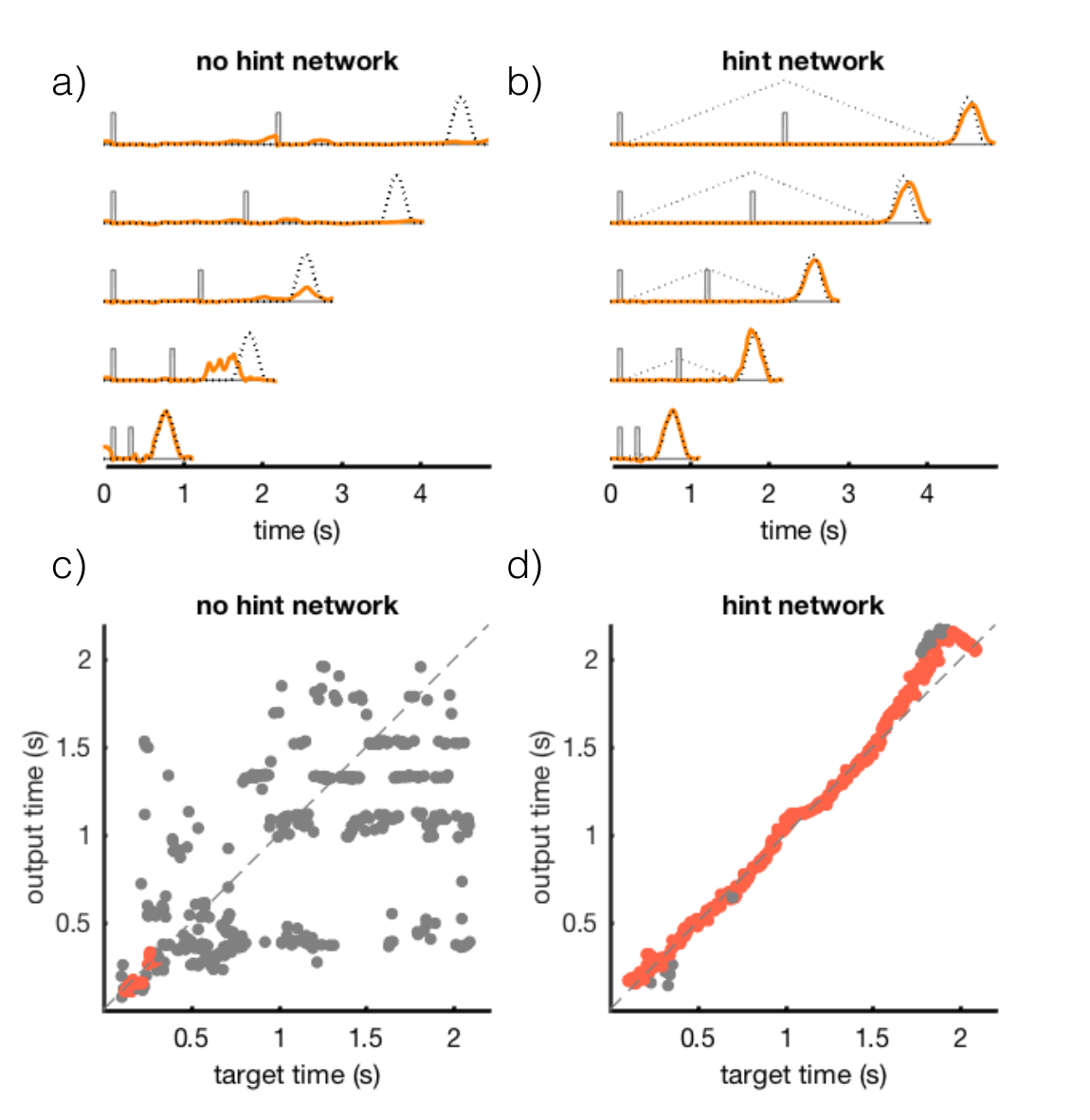}}
	\caption{Performance results for networks trained on the interval timing task. ({\bf a \& \bf b}) $f_{\inMy}(t)$ (grey), $f_{\hintMy}(t)$ (grey dotted), $f_{\outMy}(t)$ (black dotted) and $z(t)$ (orange) for a network of 1000 units. Networks trained with full-FORCE learning without ({\bf a}) and with ({\bf b}) a hint for various interpulse intervals (100, 600, 1100, 1600 and 2100 ms from bottom to top). ({\bf c \& \bf d}) Target response time plotted against the generated response time without ({\bf c}) and with ({\bf d}) hints. Each dot represents the timing of the peak of the network output response on a single test trial. Grey dots indicate that the network output did not meet the conditions to be considered a ``correct'' trial (see main text). Red dots show correct trials.}
	
	\label{fig:Figure_7}

\end{figure}

As our first example, we present a network performing an interval matching task~(\hyperref[fig:Figure_7]{Figure~\ref{fig:Figure_7}}) inspired by the ``ready-set-go'' task investigated by~\citet{Jazayeri_Shadlen_2010}.  In this task, the network receives a pair of input pulses $f_{\inMy}(t)$ of amplitude 1.0 and duration 50 ms. The job of the network is to generate an output response at a delay after the second pulse equal to the interval between the first and second pulses. The time interval separating the two input pulses is selected at random on each trial from a uniform distribution between 0.1 and 2.1 seconds. Following the input pulses, the network must respond with an output pulse $f_{\outMy}(t)$ that is a smooth, bump-like function (generated from a beta distribution with parameters $\alpha = \beta = 4$) of duration 500 ms and peak amplitude 1.5. Trials begin at random times, with an inter-trial interval drawn from an exponential distribution with a mean of 2.4 seconds. This task is difficult for any form on recurrent network learning because of the long (in terms of the 10 ms network time constant) waiting period during which no external signal is received by the network.

To assess the performance of a trained network, we classified test trials as ``correct'' if the network output matched the target output with a normalized error less than 0.25 within a window of 250 ms around the correct output time. A network of 1000 neurons, trained by full-FORCE without using a hint performs this task poorly, only correctly responding on 4\% of test trials (\hyperref[fig:Figure_7]{Figure~\ref{fig:Figure_7}a)}. The network responded correctly only on those trials for which the interpulse interval was short, less than 300 ms (\hyperref[fig:Figure_7]{Figure~\ref{fig:Figure_7}c)}.  This is because, when the desired output pulse $f_{\outMy}(t)$ is the only input provided, the dynamics of the target-generating network during the waiting period for long intervals are insufficient to maintain timing information. Therefore, we added a hint input to the target-generating network to makes its dynamics more appropriate for solving the task. Specifically, we chose $f_{\hintMy}$ to be a function that ramped linearly in time between the first and second pulses and then decreased linearly at the same rate following the second pulse (\hyperref[fig:Figure_7]{Figure~\ref{fig:Figure_7}b, grey dotted line)}. The rate of increase is the same for all trials.  A temporal signal of this form should be useful because it ensures that the task-performing network encodes both the elapsed time during the delay and the time when the waiting period is over (indicated by its second zero crossing).  

Providing a hint improves performance significantly.  An identical network trained with full-FORCE when this hint is included correctly responded to 91\% of test trials (\hyperref[fig:Figure_7]{Figure~\ref{fig:Figure_7}b}). The bulk of the errors made by the network trained with hints were modest timing errors, predominantly occurring on trials with very long delays, as can be seen in \hyperref[fig:Figure_7]{Figure~\ref{fig:Figure_7}d}. These results illustrate the power of using task hints in our network training framework: including a hint allowed a network to transition from performing the task extremely poorly to performing almost perfectly.

\subsec{Delayed comparison task}

In the previous example, we illustrated the utility of hints on a task that could not be trained without using one.  However, not all tasks require hints to be performed successfully.  Is there any benefit to include hints for tasks that can be performed without them? Here we train networks on a task that can be performed reasonable well without hints and show that only a network trained with a hint can extrapolate beyond the domain of the training data. Extrapolation is an important feature because it indicates that a more general, and therefore potentially more robust, task solution has been constructed.

\begin{figure}[h!]
	\centerline {\includegraphics[width=4in]{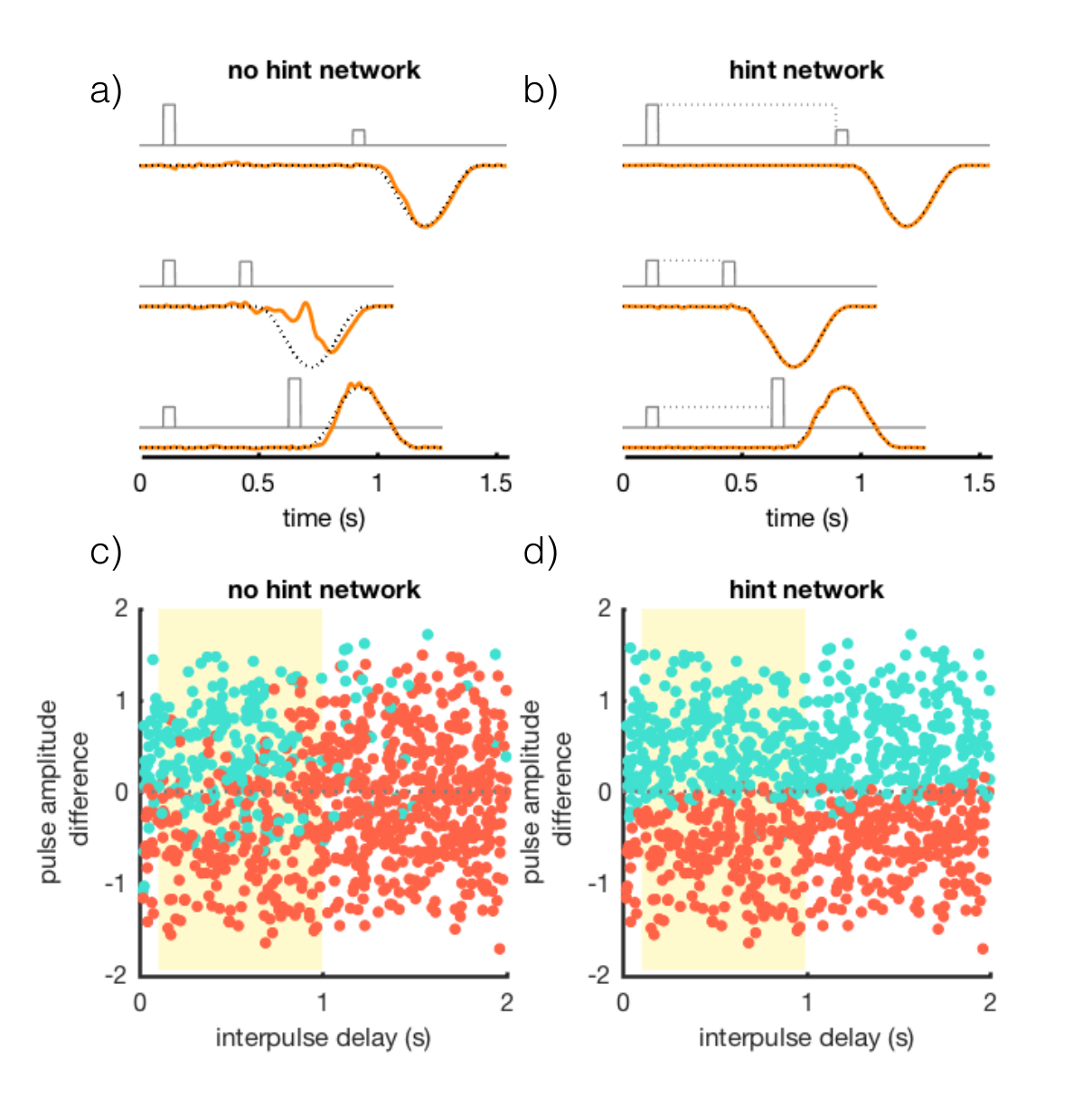}}
	\caption{Performance results for networks trained on the delayed comparison task. ({\bf a \& \bf b}) $f_{\inMy}(t)$ (grey), $f_{\hintMy}(t)$ (grey dotted), $f_{\outMy}(t)$ (black dotted) and $z(t)$ (orange) for a network of 1000 units. Networks trained with full-FORCE learning without ({\bf a}) and with ({\bf b}) hints. Three different trials are shown from bottom to top: an easy ``+'' trial, a difficult ``-'' trial, and an easy ``-'' trial. ({\bf c \& \bf d}) Test performance for networks trained without ({\bf c}) and with ({\bf d}) a hint. Each dot indicates a test trial and the dot color indicates the reported output class (``+'' cyan or ``-'' red). The horizontal axis is the interpulse delay and the yellow region indicates the training domain. The vertical axis indicates the pulse amplitude difference.}
	
	\label{fig:Figure_8}

\end{figure} 

We present a version of a delayed comparison task that has been studied extensively by both the experimental and theoretical neuroscience communities~\citep{Romo,MachensBrody,Barak_Abbott}. In this task, a network receives a pair of input pulses $f_{\inMy}(t)$ of duration 50 ms. Each pulse has an amplitude that varies randomly from trial to trial, selected from a uniform distribution between 0.125 and 1.875. Furthermore, during training the time interval separating the two input pulses is selected at random from a uniform distribution between 0.1 and 1 second, thereby preventing the network from being able to anticipate the timing of the second pulse. Following the pulses, the network must respond with an output pulse $f_{\outMy}(t)$ defined as for the interval timing task, but positive if the difference in magnitude of the two input pulses is positive (``+'' trials) and negative if the difference was negative (``-'' trials)~(\hyperref[fig:Figure_8]{Figure~\ref{fig:Figure_8}a \& b}). Trials on which the input pulse height difference is small are clearly difficult because smalls errors will cause the network to misclassify the input. As in the interval timing task, trials begin at random times with inter-trial intervals of random duration.

To assess the performance of a trained network, we classified test trials as ``correct'' if the network output matched the target output with a normalized error less than 0.25 and ``incorrect'' if the network output matched the \emph{opposite} class target output with a normalized error less than 0.25. Trials where the network output did not meet either of these criteria were classified as undertermined and were not included in test performance. Networks trained without a hint correctly classified 71\% of test trials. Three example trials are shown in~\hyperref[fig:Figure_8]{Figure~\ref{fig:Figure_8}a} and a summary of test performance can be seen in~\hyperref[fig:Figure_8]{Figure~\ref{fig:Figure_8}c}.

Next we trained an identical network with a hint to see if doing so could increase and extend task performance. The hint signal for this task was a constant input equal to the magnitude of the first pulse that lasted for the entire interpulse interval. A hint of this form has the effect of driving the network dynamics to an input specific fixed-point, thereby constructing a memory signal of the value of the first pulse. Including a task hint increased performance on this task, where the hint trained network could correctly classify 95\% of test trials~(\hyperref[fig:Figure_8]{Figure~\ref{fig:Figure_8}b}).  More importantly, it allowed the network to extrapolate.

To examine the ability of both trained networks to extrapolate beyond the range of training data, we tested performance with interpulse delays selected uniformly between 20 ms and 2 seconds. Intervals less than 100 ms and greater than 1 second were never seen by the network during training. The network trained without a task hint performed just above chance on these trials, correctly classifying only 53\% of trials. The performance of the network trained with the hint, on the other hand, was hardly diminished when compared to its performance over the range for which it was trained; 94\% compared to 95\% over the training range (\hyperref[fig:Figure_8]{Figure~\ref{fig:Figure_8}d}). These results illustrate the potential for task hints to simplify the dynamics used by a network to solve a task, thereby allowing a more general solution to emerge, one that supports extrapolation beyond the training data.

\sec{Discussion}

We have presented a new target-based method for training recurrently connected neural networks. Because this method modifies the full recurrent connectivity, the networks it constructs can perform complex tasks with a small number of units and considerable noise robustness.  The speed and simplicity of the recursive least-squares algorithm makes this an attractive approach to network training.  In addition, by dividing the problem of target generation and task performance across two separate networks, full-FORCE provides a straightforward way to introduce additional input signals that act as hints during training. Utilizing hints can improve post-training network performance by assuring that the dynamics are appropriate for the task. Like an animal or human subject, a network can fail to perform a task either because it is truly incapable of meeting its demands, or because it fails to ``understand'' the nature of the task. In the former case, nothing can be done, but in the latter the use of a well-chosen hint during training can improve performance dramatically.  Indeed, we have used full-FORCE with hints to train networks to perform challenging tasks that we previously had considered beyond the range of the recurrent networks we use. 

The dominant method for training recurrent neural networks in the machine learning community is backpropagation through time \citep{Werbos}.  Backpropagation-based learning has been used successfully on a wide range of challenging tasks including language modeling, speech recognition and handwriting generation~\citep[][and references therein]{Lecun2015} but these tasks differ from the interval timing and delayed comparison tasks we investigate in that they do not have long periods during which the input to the network is silent.  This lack of input requires the network dynamics alone to provide a memory trace for task completion, precisely the function of our hint signal. When using backpropagation through time, the network needs to be unrolled a sufficient number of time steps to capture long time-scale dependencies, which, for our tasks, would mean unrolling the network at least the maximum length of the silent periods, or greater than 2,000 time steps. Given well-known gradient instability problems~\citep{Bengio1994}, this is a challenging number of unrollings. While we make no claim that gradient-based learning is unable to learn tasks such as those we study, full-FORCE's success when a well-selected hint is used is striking given the algorithmic simplicity of RLS and architectural simplicity of the units we use.  Although we have highlighted the use of hints in the full-FORCE approach, it is worth noting that hints of the same form can be used in other types of learning including traditional FORCE learning and backpropagation-based learning~\citep{Abu-Mostafa1994,Abu-Mostafa1995}. In these cases, $f_{\hintMy}(t)$ would be added as a second target output, along with $f_{\outMy}(t)$. By requiring the learning procedure to match this extra output, the network dynamics can be shaped appropriately. Within backpropagation-based learning, doing so might avoid gradient instability problems and perhaps release these methods from a reliance on more complex networks units such as LSTMs~\citep[Long Short-Term Memory units;][]{Hochreiter2}. 
 


Studies of recurrent neural networks that perform tasks vary considerably in their degree of ``design".  At one end of this spectrum are hand-tuned models in which a particular solution is hard-wired into the network by construction \citep[examples of this include][]{Seung,MachensBrody}.  Networks of this type have the advantage of being easy to understand, but their dynamics tends to be simpler than what is seen in the activity of real neurons \citep{MillerEtAl,Barak_Abbott}.  At the other extreme are networks that rely on dynamics close to the chaos of a randomly connected network \citep{Sompolinsky_Crisanti_Sommers_1988}, a category that includes the work of  \citet{Laje_Buonomano_2013} and much of the work in FORCE and echo-state networks.  This approach is useful because it constructs a solution for a task in a fairly unbiased way, which can lead to new insights.  Networks of this type tend to exhibit fairly complex dynamics, an interesting feature~\citep{SussilloBarak} but, from a neuroscience perspective, this complexity can exceed that of the data~\citep{Barak_Abbott,SussilloChurchland}.  This has led some researchers to resort to gradient-based methods where regularizers can be included to reduce dynamic complexity to more realistic levels~\citep{SussilloChurchland}.  Another approach is to use experimental data directly in the construction of the model~\citep{Goldman,Rajan_Harvey_Tank_2016}.  Regularization and adherence to the data can be achieved in full-FORCE through the use of well-chosen hints imposed on the target-generating network.  More generally, hints can be used for a range of purposes, from imposing a design on the trained network to regulating its dynamics. Thus, they supply a method for controlling where a model lies on the designed-to-random spectrum.

By splitting the requirements of target generation and task performance across two networks, full-FORCE introduces a freedom that we have not fully exploited.  The dimension of the dynamics of a typical recurrent network is considerably less than its size $N$ \citep{Rajan2010}. This implies that the dynamics of the target-generating network can be described by a smaller number of factors such as principle components, and these factors, rather than the full activity of the target-generating network, can be used to produce targets, for example by linear combination. Furthermore, in the examples we provided, the target-generating network was similar  to the task-performing network in structure, parameters ($\tau$ = 10 ms for both) and size.  We also made a standard choice for $\JD$ to construct a randomly connected target-generating network.  None of these choices is mandatory, and a more creative approach to the construction of the target-generating network (different  $\tau$, different size, modified $\JD$, etc.) could enhance the properties and performance of networks trained by full-FORCE, as well as extending the range of tasks that they can perform.  

\subsec{Acknowledgments}

We thank Raoul-Martin Memmesheimer for fruitful discussions regarding an earlier version of this work. We thank Ben Dongsung Huh for suggesting the name full-FORCE that integrates the original FORCE acronym and David Sussillo, Omri Barak and Vishwa Goudar for helpful suggestions and conversations. Research supported by NIH grant MH093338, the Gatsby Charitable Foundation and the Simons Foundation. BD supported by a NSF Graduate Research Fellowship.

\end{document}